\pdfoutput=1 
\documentclass{article}

\usepackage{authblk}
\usepackage{blindtext}
\usepackage{multicol}
\usepackage{xcolor}
\usepackage{amssymb}
\usepackage{textcomp}
\usepackage{amsthm}
\usepackage{caption}
\usepackage{subfig}
\usepackage{makecell}
\usepackage{algorithm}
\usepackage{algorithmic}
\usepackage{stackengine}
\usepackage{booktabs}
\usepackage{url}
\usepackage{booktabs}

\usepackage{etoolbox}
\usepackage{graphicx}
\usepackage{amsmath}
\usepackage{tikz}
\usetikzlibrary{positioning}
\newcommand{\RomanNumeralCaps}[1]
    {\MakeUppercase{\romannumeral #1}}
\newcommand\mydots{\ifmmode\ldots\else\makebox[1em][c]{.\hfil.\hfil.}\thinspace\fi}
\newcommand{\argmin}[1]{\underset{#1}{\operatorname{argmin}}}

\usepackage[left=2cm, right=2cm, top=3cm, bottom=3cm]{geometry}

\title{Tensor Network Kalman Filtering for Large-Scale LS-SVMs}
\author[1]{Maximilian Lucassen}
\author[2]{Johan A.K. Suykens}
\author[1]{Kim Batselier}
\affil[1]{Delft Center for Sytems and Control, Delft University of Technology, Mekelweg 2, 2628 CD, Delft, The Netherlands.}
\affil[2]{Department of Electrical Engineering,  ESAT-STADIUS, Katholieke Universiteit Leuven, Kasteelpark Arenberg 10, Leuven, Belgium.}
\date{}

\date{\empty}

\begin{document}

\maketitle

\begin{abstract}
Least squares support vector machines are a commonly used supervised learning method for nonlinear regression and classification. They can be implemented in either their primal or dual form. The latter requires solving a linear system, which can be advantageous as an explicit mapping of the data to a possibly infinite-dimensional feature space is avoided. However, for large-scale applications, current low-rank approximation methods can perform inadequately. For example, current methods are probabilistic due to their sampling procedures, and/or suffer from a poor trade-off between the ranks and approximation power. In this paper, a recursive Bayesian filtering framework based on tensor networks and the Kalman filter is presented to alleviate the demanding memory and computational complexities associated with solving large-scale dual problems. The proposed method is iterative, does not require explicit storage of the kernel matrix, and allows the formulation of early stopping conditions. Additionally, the framework yields confidence estimates of obtained models, unlike alternative methods. The performance is tested on two regression and three classification experiments, and compared to the Nystr\"om and fixed size LS-SVM methods. Results show that our method can achieve high performance and is particularly useful when alternative methods are computationally infeasible due to a slowly decaying kernel matrix spectrum.
\end{abstract}

\section{Introduction}

Kernel methods are a class of nonlinear modeling methods based on mapping the nonlinear input data to a high-dimensional feature space. In this high-dimensional feature space, the inner product of the data is implicitly computed by a user-defined kernel function, correlating the data points through their similarity. Due to their formulations, the data can be modeled with the simplicity of linear algorithms \cite{Vapnik1995} \cite{LearningWithKernels} \cite{KernelMethodPatternAnalysis}. As a result, kernel methods are an attractive modeling class because of the insight and ease they provide, and are widely applied in numerous disciplines such as control \cite{KocijanGP}, machine learning \cite{FoundationsOfMachineLearning} and signal processing \cite{KernelMethodsSignalProcessing}. 
    
The SVM theory was initially proposed by Vapnik \cite{Vapnik1995}, in which a cost function is minimized by solving either the parametric primal problem or the non-parametric dual problem. Either case, the modeling is solved with quadratic programming, generating a unique global solution. In the least squares SVM (LS-SVM) one reformulates the original SVM problem by changing the loss function to a least squares form, and altering the inequality constraints to equality constraints \cite{LS-SVM} \cite{LSSVM-classifiers} \cite{ApproximateConfPredInt}. This makes the resulting LS-SVM dual problem much easier to solve compared to its SVM counterpart.
     
\begin{figure}[h!]
     \centering
     \includegraphics[scale=0.25]{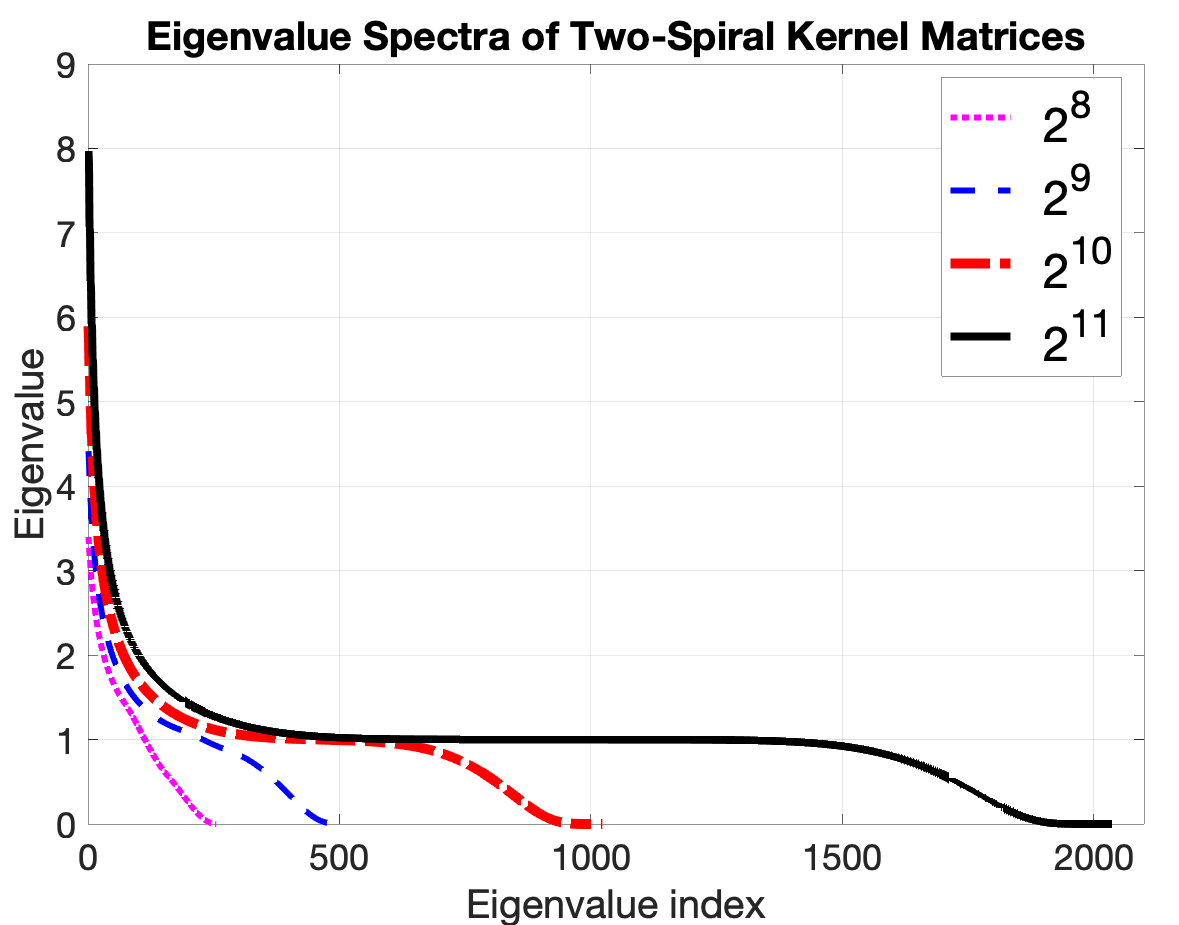}
     \caption{Eigenvalue spectra of differently sized two-spiral kernel matrices. The eigenvalues only become insignificant at the final indices. These spectra are difficult to approximate with low-rank methods that employ eigendecompositions as many eigencomponents and samples need to be used for an accurate solution.}
  \label{fig:EigSpecTwoSpiral}
\end{figure}

The computational complexity of solving the LS-SVM dual problem is $O(N^3)$ for direct methods and $O(N^2L)$ for iterative methods, where $N$ denotes the number of data points and $L$ the number of iterations. For small and medium-scale problems, general solvers from the direct and iterative methods classes are applicable \cite{BHamers} \cite{MatrixComputations}. However, these methods cannot be used due to memory and computational requirements when the number of data points $N$ grows exponentially large. By this we mean that $N$ can be written as $n^d$ for some $n$ and $d$. As the exponent $d$ grows, more and more methods will become problematic to use. This exponential growth is an instance of the "curse of dimensionality". One approach of solving large dual problems is to use a low-rank approximation of the kernel matrix, which is the cornerstone for the Nystr\"{o}m \cite{WilliamsSeeger2001} and fixed-size LS-SVM (FS-LSSVM) methods~\cite{OptimizedFSLSSVM}. These methods aim to work around the prohibitive scaling by training on a subset of the data and by employing the primal-dual relationship of the LS-SVM. Two drawbacks to current low-rank approximation methods exist. Firstly, no confidence bounds of the approximate solution are obtained, only theoretical statistical guarantees can be given. Also, the size of the sample subset significantly impacts the performance in accuracy and the memory and computational complexity. Methods that rely on a low-rank approximation of the kernel matrix inherently depend on a kernel spectrum that exhibits a strong decay. When such a decay is lacking then these methods are expected to perform poorly. An example of such a slowly decaying spectrum is shown in Fig~\ref{fig:EigSpecTwoSpiral} for the two-spiral problem. As the sample size increases one can see that the spectrum exhibits an ever-increasing plateau of important eigenvalues. This observation motivates the development of our proposed recursive method that relies on a low-rank tensor decompositions instead. The power of our proposed method is demonstrated in the experiments where we train an LS-SVM for the two-spiral problem on $2^{20}$ training points and achieve a 100\% accuracy during validation.

 \subsection{Contributions of This Work}
In this paper, the curse of dimensionality associated with solving the large-scale LS-SVM dual problem and the inability of current low-rank approximation methods to handle slow-decaying kernel spectra are addressed. The dual problem is cast into a tensor train (TT) form, which reduces the storage complexity from $O(n^d)$ down to $O(dnr^2)$, where $r$ denotes the TT-rank. In this form, the dual matrix never has to be constructed explicitly, nor any other matrix. A recursive Bayesian procedure is then developed to solve the large-scale dual problem in TT form by implementation of a tensor network Kalman filter (TNKF). Such a Bayesian approach also results in a covariance matrix of the dual variables, which can be used to implement an early-stopping criterion. The contributions of this work are:
     
\begin{itemize}
     \item Circumventing the curse of dimensionality of solving large-scale LS-SVM dual problems by using a low-rank TT representation. 
     \item A non-sampling based algorithm that evaluates the entire dual matrix and constructs low-rank tensor network approximations of each row. 
     \item Implementation of a recursive Bayesian filter that, unlike current low-rank approximation methods, allows the computation of confidence bounds on predictions. 
\end{itemize}     
     
 \subsection{Related Work}

Many direct and iterative methods such as the LU decomposition and conjugate gradient \cite{MatrixComputations} \cite{BHamersIterativeMethods} are available for learning small-scale LS-SVMs. The $\mathcal{O}(n^{2d})$ storage complexity and $\mathcal{O}(n^{3d})$ computational complexity of direct methods inhibit their application for large-scale problems. Iterative solvers still suffer from an exponential complexity $\mathcal{O}(n^{2d}l)$, where $l$ denotes the number of iterations. A common way to deal with this difficulty is to employ low-rank approximation methods by utilizing a subset of $S$ samples to approximate the solution. A trade-off between the obtainable accuracy, and the associated computational and memory requirements is made for the problem to be feasible. Two well-known low-rank approximation techniques for LS-SVMs are the Nystr\"om method~\cite{WilliamsSeeger2001} and FS-LSSVM~\cite{OptimizedFSLSSVM}. Both are sampling-based methods that rely on the eigendecomposition of the kernel matrix. The Nystr\"om method solves the LS-SVM dual problem with computational and storage complexities of $\mathcal{O}(S^2n^{d}+S^3)$ and $\mathcal{O}(Sn^{d})$, respectively. FS-LSSVM estimates a nonlinear mapping function in the dual and then solves the primal problem. The computational and storage complexities of FS-LSSVM are $\mathcal{O}(S^2n^{d}+2S^3)$ and $\mathcal{O}(S^2)$ \cite{BHamers}, respectively. For both methods, the sampling procedure has a significant impact on the performance. Many sampling procedures exist such as uniform (random) sampling, or if computationally achievable, more advanced adaptive techniques can be used, such as sparse greedy matrix approximation \cite{SGMA}, leverage-scores \cite{LeverageScores}, column-norm sampling \cite{ColumnNormSampling}, and Renyi-entropy sampling \cite{LS-SVM}. Other approaches to deal with large-scale linear systems commonly focus on parallel computation or sparsity. For example, in \cite{ma2018kernel} a parallel GPU implementation is used for kernel machines. The linear scaling associated with increasing batch sizes is extended with adaptive kernels to speed up training times and more efficient parallelization. In \cite{ChenZ}, the computational complexity to train polynomial classifiers are significantly reduced through parallel implementations of tensor trains. In \cite{Sparsity}, a non-iterative pruning method is employed to generate a sparse LS-SVM solution. By using globally representative points a support vector subset is found with lower computational complexities than traditional iterative pruning methods.

This paper is organized as follows. In Section \RomanNumeralCaps{2} the basics of LS-SVMs are covered. The Kalman filter is also introduced in context of Bayesian filtering. In Section \RomanNumeralCaps{3}, a brief introduction to tensor networks is given, including required operations and the TT decomposition for the TNKF algorithm. The final TNKF algorithm is presented in Section \RomanNumeralCaps{4} and Section \RomanNumeralCaps{5}, covering its derivation, implementation and complexities. The TNKF is analyzed on five datasets in Section \RomanNumeralCaps{6}. Lastly, conclusions and further work are discussed in Section \RomanNumeralCaps{7}.

The notation and abbreviations used in this paper are given in Table~\ref{tab:Notation} and Table~\ref{tab:Abbreviations}, respectively. 

\begin{table}[h!]
 \caption{Used Notation}
 \label{tab:Notation}
 \centering
 \begin{tabular}{|l|l|}
    \hline
    Scalars               & (a,b,...) \\ \hline
    Vectors     & ($\boldsymbol{a}$, $\boldsymbol{b}$,...)  \\ \hline
    Matrices            & ($\boldsymbol{A}$, $\boldsymbol{B}$,...) \\ \hline
    Tensors       & ($\mathcal{A}$,$\mathcal{B}$,...) \\ \hline
    Tensor Train of $\mathcal{A}$   & TT$(\mathcal{A})$ \\ \hline
    Matrix transpose   &        ($\boldsymbol{a}^{\intercal}$,$\boldsymbol{A}^{\intercal}$,...) \\ \hline
    Identity matrix of size $n$    &    $\boldsymbol{I}_{n}$ \\ \hline
    Kernel matrix of size $n$ by $n$       &   $\boldsymbol{\Omega}_{n}$ \\ \hline
 \end{tabular}
\end{table}

\begin{table}[h!]
 \caption{Used Abbreviations}
 \label{tab:Abbreviations}
 \centering
 \begin{tabular}{|l|l|}
    \hline
    TT      & Tensor train \\ \hline
    SVM     & Support-vector-machines\\ \hline
    LS-SVM  & Least squares support-vector-machines\\ \hline
    FS-LSSVM & Fixed size least squares support-vector-machines\\\hline
    KKT     & Karush-Kuhn-Tucker  \\ \hline
    SVD & Singular value decomposition \\ \hline
    TNKF    & Tensor network Kalman filter\\
    \hline
 \end{tabular}
\end{table}

\section{Solving LS-SVMs with Bayesian filtering}

\subsection{Least Squares Support Vector Machines}
In what follows, we consider the regression problem. The development of our method is easily adjusted to the classification problem. Consider a training set $\{\boldsymbol{x}_k,y_k\}_{k=1}^{n^d}$ of inputs $\boldsymbol{x}_k$ $\in$ $\mathbb{R}^f$ and outputs $y_k$ $\in$ $\mathbb{R}$. The regression model in the primal space is
\begin{align*}
    y_k &= \varphi(\boldsymbol{x}_k)^T\, \boldsymbol{w} + b + e_k,
\end{align*}
where $b$ denotes the bias, $\varphi(\cdot): \mathbb{R}^f \rightarrow \mathbb{R}^{f_h}$ denotes a feature map to a possibly infinite-dimensional space and $e_k$ are the residuals. The LS-SVM primal problem is then to estimate the weights $\boldsymbol{w} \in \mathbb{R}^{f_h}$, bias $b$ and residuals $e_k$ from
\begin{align*}
    \argmin{\boldsymbol{w},b,e}\; \frac{1}{2} \boldsymbol{w}^T \boldsymbol{w} &+ \frac{\gamma}{2}\, \sum_{k=1}^{n^d} e_k^2 \\
    \textrm{such that } \; y_k = \varphi(\boldsymbol{x}_k)^T\, \boldsymbol{w} + b &+ e_k, k=1,\ldots,n^d.
\end{align*}
Solving the primal problem can be advantageous when \mbox{$n^d \gg f_h$}. However, the primal problem is often impossible because $\varphi(\cdot)$ is usually unknown, difficult to determine and possibly infinite-dimensional. The dual problem circumvents this problem by rewriting the primal problem in terms of a kernel function $K(\boldsymbol{x}_i,\boldsymbol{x}_j)=\varphi(\boldsymbol{x}_i)^T\varphi(\boldsymbol{x}_j)$ as 
\vspace{0.1cm}
\begin{equation}    
    \label{LSSVM_regression:dual}
             \begin{bmatrix} 
                        0 & \boldsymbol{1}^{\intercal}\\
    \boldsymbol{1} &  \boldsymbol{\Omega}_{n^d} + \boldsymbol{I}_{n^d}/\gamma\\         
    \end{bmatrix}
    \begin{bmatrix}
        b \\
        \boldsymbol{\alpha}
    \end{bmatrix}
    =
    \begin{bmatrix}
        0 \\
        \boldsymbol{y}
    \end{bmatrix}.\\ 
\end{equation}
The $n^d \times n^d$ matrix $\boldsymbol{\Omega}_{n^d}$ is called the kernel matrix and is defined as $\Omega(i,j) = K(\boldsymbol{x}_i,\boldsymbol{x}_j)$. The predictor can then also be written in terms of the dual variables $\boldsymbol{\alpha} \in \mathbb{R}^{n^d}$ as
\vspace{0.1cm}
\begin{equation}
    y(\boldsymbol{x})=\sum_{k=1}^{n^d} \alpha_{k}\; K\left(\boldsymbol{x}, \boldsymbol{x}_{k}\right)+b.
\label{eq:dualmodel}
\end{equation}

The dual variables $\boldsymbol{\alpha}$ are related to the residuals of the primal problem $e_k$ through $\alpha_k = \gamma e_k$. More detailed derivations of both the primal and dual problem can be found in~\cite{LS-SVM}.

\subsection{Bayesian Filtering and the Kalman Filter}
\label{sec:BayesianFilter}
A recursive Bayesian filtering approach is adopted to solve the LS-SVM dual problem. Such a recursive approach makes it possible to quantify the uncertainty of our solution during the recursion, which enables the construction of an early-stopping criterion such that large-scale datasets can be trained. Here, the dual problem~\eqref{LSSVM_regression:dual} is written as
\begin{align*}
\boldsymbol{C} \bar{\boldsymbol{\alpha}} + \boldsymbol{r} = \bar{\boldsymbol{y}}
\end{align*}
where $\boldsymbol{C}$ is the dual problem matrix, $\bar{\boldsymbol{\alpha}} = \begin{pmatrix} b & \boldsymbol{\alpha}^T \end{pmatrix}^T$, \mbox{$\bar{\boldsymbol{y}} = \begin{pmatrix}0&\boldsymbol{y}^T \end{pmatrix}^T$} and the vector $\boldsymbol{r}$ contains the residuals of the dual problem. The goal now is to compute the posterior $\mathcal{P}(\bar{\boldsymbol{\alpha}}|\bar{y}_{1:n^d})$ of $\bar{\boldsymbol{\alpha}}$ conditioned on the data $\bar{y}_{1:T}$. Using Bayes' rule this posterior can be written as
\begin{equation}
    \mathcal{P}\left( \bar{\boldsymbol{\alpha}} \mid \bar{y}_{1:n^d} \right) = \frac{\mathcal{P}\left( \bar{y}_{1:n^d}  \mid \bar{\boldsymbol{\alpha}} \right) \mathcal{P}\left( \bar{\boldsymbol{\alpha}} \right)}{\mathcal{P}\left( \bar{y}_{1:n^d} \right)},
\end{equation}
where $\mathcal{P}\left( \bar{y}_{1:n^d}  \mid \bar{\boldsymbol{\alpha}} \right)$ is called the likelihood and $\mathcal{P}\left( \bar{\boldsymbol{\alpha}} \right)$ the prior.  

We assume that the primal residuals $e_k$ are zero-mean Gaussian random variables $\boldsymbol{e} \sim \mathcal{N}(0,\sigma_e^2 \boldsymbol{I})$. From $\alpha_k = \gamma e_k$ it then follows that the prior on ${\bar{\boldsymbol{\alpha}}}$ is also normally distributed $\bar{\boldsymbol{\alpha}} \sim \mathcal{N}(0,\gamma^2 \sigma_{e}^2 \boldsymbol{I})$. For convenience we define the short-hand notation $\boldsymbol{P_0} :=\gamma^2 \sigma_{e}^2 \boldsymbol{I}$. The regularization parameter $\gamma$ hence reflects our confidence in the prior. The residuals $\boldsymbol{r}$ of the dual problem are also assumed to be normally distributed $\boldsymbol{r} \sim \mathcal{N}(0,\sigma_{r}^2\boldsymbol{I})$. Because the prior and the likelihood are Gaussian, the posterior $\mathcal{P}(\bar{\boldsymbol{\alpha}}|\bar{y}_{1:T})$ is also Gaussian. From Bayes' rule an analytic solution can be derived to calculate the posterior \mbox{$\mathcal{P}(\bar{\boldsymbol{\alpha}}|\bar{y}_{1:n^d}) = \mathcal{N}(\boldsymbol{m}_{n^d}, \boldsymbol{P}_{n^d})$} with
\begin{equation}
\label{inverseBayesian}
\begin{split}
\boldsymbol{m}_{n^d} &=\left[\boldsymbol{P}_{0}^{-1}+\frac{1}{\sigma_{r}^2} \boldsymbol{C}^{\intercal} \boldsymbol{C}\right]^{-1}\left[\frac{1}{\sigma_{r}^2} \boldsymbol{C}^{\intercal} \bar{\boldsymbol{y}}\right], \\
\boldsymbol{P}_{n^d} &=\left[\boldsymbol{P}_{0}^{-1}+\frac{1}{\sigma_{r}^2} \boldsymbol{C}^{\intercal} \boldsymbol{C}\right]^{-1}.\\
\end{split}
\end{equation}
The matrix inverse computation has a prohibitive computational complexity of $\mathcal{O}(n^{3d})$. By realizing that posterior distributions can also serve as prior distribution for each succeeding observation, a recursive framework can be developed that avoids the explicit matrix inversion. In this article, single observations are used to update the $\bar{\boldsymbol{\alpha}} $ distribution, which is equivalent to solving the dual problem row-by-row. The following model is used to describe the solution for $\bar{\boldsymbol{\alpha}}$ at iteration $k$,
\begin{align}
    \bar{\boldsymbol{\alpha}}_{k+1} &= \bar{\boldsymbol{\alpha}}_{k} + \boldsymbol{q}\\ 
    \bar{y}_{k} &= \boldsymbol{c}_k^T\, \bar{\boldsymbol{\alpha}}_{k}+r_k
\label{eq:StateSpace_regression}
\end{align}

where $\boldsymbol{c}_k^T$ is the $k$th row of the $\boldsymbol{C}$ matrix and $\boldsymbol{q}$ is a stochastic term that allows us to introduce a forgetting factor $\lambda$ that weights past measurements. We also assume that $\boldsymbol{q}$ is normally distributed $\boldsymbol{q} \sim \mathcal{N}(\boldsymbol{0}, \boldsymbol{Q})$. Note that $\bar{\boldsymbol{\alpha}}_k$ denotes the estimate of $\bar{\boldsymbol{\alpha}}$ at iteration $k$ and not the $k$th component of the vector $\bar{\boldsymbol{\alpha}}$. Assuming the Markov property
\begin{align*}
    \mathcal{P}\left( \bar{\boldsymbol{\alpha}}_k | \bar{\boldsymbol{\alpha}}_{1:k-1}, \bar{y}_{1:k-1} \right) =  \mathcal{P}\left( \bar{\boldsymbol{\alpha}}_k | \bar{\boldsymbol{\alpha}}_{k-1} \right)
\end{align*}
and the conditional independence of the measurements
\begin{align*}
    \mathcal{P}\left( \bar{y}_k | \bar{\boldsymbol{\alpha}}_{1:k}, \bar{y}_{1:k-1} \right) =  \mathcal{P}\left( \bar{y}_k | \bar{\boldsymbol{\alpha}}_{k} \right)
\end{align*}
then the posterior $\mathcal{P}\left(\bar{\boldsymbol{\alpha}}_k \mid \bar{y}_{1: k}\right)$ can be written as
\begin{equation}
    \label{BayesRecursive}
    \mathcal{P}\left(\bar{\boldsymbol{\alpha}}_k \mid \bar{y}_{1: k}\right) \propto
    \mathcal{P}\left(\bar{\boldsymbol{\alpha}}_{k-1} \mid \bar{y}_{1: k-1}\right)
    \mathcal{P}\left(\bar{y}_{k} \mid \bar{\boldsymbol{\alpha}}_{k} \right). 
\end{equation}
This recursive Bayesian framework directly leads to the Kalman filter equations, an optimal closed-form algorithm for linear stochastic models \cite{BayesianFilteringAndSmoothing} \cite{verhaegen_verdult_2007}. The Kalman filter equations tells us how to compute the mean vector $\boldsymbol{m}_k$ $\in$ $\mathbb{R}^{n^d}$ and covariance matrix $\boldsymbol{P}_k$ $\in$ $\mathbb{R}^{n^d\times n^d}$ of the posterior $\mathcal{P}\left(\bar{\boldsymbol{\alpha}}_k \mid \bar{y}_{1: k}\right)$ at iteration $k$
\begin{align}
\begin{split} \label{eq:Prediction_Step_KF}
    \boldsymbol{m}_{k}^{-} &= \boldsymbol{m}_{k-1} \\
    \boldsymbol{P}_{k}^{-} &= \boldsymbol{P}_{k-1} + \boldsymbol{Q}\\
    \end{split}\\
    \begin{split}
    \label{eq:Measurement_Step_KF}
        v_k &= \bar{y}_k - \boldsymbol{c}_k\boldsymbol{m}_{k}^{-} \hspace{0.5cm}\\ 
    s_k &= \boldsymbol{c}_{k}\boldsymbol{P}_{k}^{-}\boldsymbol{c}_k^{\intercal}+\sigma_r^2 \\
    \boldsymbol{k}_k &= \boldsymbol{P}_{k}^{-}\boldsymbol{c}_k^{\intercal}s_k^{-1} \\
    \boldsymbol{m}_k &= \boldsymbol{m}_{k}^{-} + \boldsymbol{k}_k v_k \\
    \boldsymbol{P}_{k} &=  \boldsymbol{P}_{k}^{-} - \boldsymbol{k}_k s_k \boldsymbol{k}_k^{\intercal}.
\end{split} 
\end{align} 
The covariance matrix $\boldsymbol{P}_k$ can be understood as a measure of uncertainty in $\bar{\boldsymbol{\alpha}}_k$. By choosing the covariance matrix of $\boldsymbol{q}$ as $\boldsymbol{Q} = (\lambda^{-1}-1) \boldsymbol{P}_{k-1}$ with $\lambda \in (0,1]$ an approximately exponentially decay weighting on past data is introduced\cite[pg.240]{ForgettingFactor}. The prediction error and its variance at iteration $k$ are $v_k$ and $s_k$, respectively. Lastly, $\boldsymbol{k}_k$ $\in$ $\mathbb{R}^{n^d}$ is the Kalman gain, which weights the observation $\bar{y}_k$ to the prior information in the computation of $\mathcal{P}\left(\bar{\boldsymbol{\alpha}}_k \mid \bar{y}_{1: k}\right)$. Termination occurs when $k=n^d$. Once the filter is finished at iteration $k$, one can set $\bar{\boldsymbol{\alpha}} = \boldsymbol{m}_k$ and compute confidence bounds from the covariance matrix $\boldsymbol{P}_{k}$, similar to Gaussian processes \cite{KocijanGP,Rasmussen06gaussianprocesses}. 

Although the $n^d \times n^d$ matrix inversion is avoided by using the Kalman equations, the explicit construction of covariance matrix $\boldsymbol{P}_k$ can still be problematic. For this reason, we propose to represent all vector and matrix quantities in the Kalman equations with tensor trains.

\section{Tensor Trains}
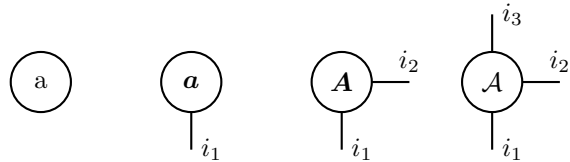
\begin{figure}[h!]
\centering
\begin{tikzpicture}
    \hspace{0.75cm}
    \draw[thick] (1,1) circle (0.4cm) node (I) {a} ;
    \draw[thick] (3,1) circle (0.4cm)   node {$\boldsymbol{a}$} ;
    \draw[thick] (3,0.6) -- (3,0.1)     node[right]{$i_1$};
    \draw[thick] (5,1) circle (0.4cm)   node {$\boldsymbol{A}$} ;
    \draw[thick] (5,0.6) -- (5,0.1)     node[right]{$i_1$};
    \draw[thick] (5.4,1) -- (5.9,1)     node[above]{$i_2$};
    \draw[thick] (7,1) circle (0.4cm)   node {$\mathcal{A}$} ;
    \draw[thick] (7,1.4) -- (7,1.9)     node[right]{$i_3$}; 
    \draw[thick] (7,0.6) -- (7,0.1)     node[right]{$i_1$};
    \draw[thick] (7.4,1) -- (7.9,1)     node[above]{$i_2$};   
\end{tikzpicture}
    \caption{Diagrammatic tensor notation of a scalar (a), vector (\textbf{a}), matrix ($\boldsymbol{A}$), and 3-dimensional tensor ($\mathcal{A}$).}
    \label{fig:VisualTensorRep}
\end{figure}

Tensors are multidimensional extensions of vectors and matrices to higher dimensions, relevant in many fields: signal processing, machine learning, and quantum physics  \cite{TensorDecSignalApp} \cite{StronglyCorrelatedSystems}. In this article, we define a tensor $\mathcal{T}$ $\in$ $\mathbb{R}^{n_1 \times n_2 ... n_d}$ as having an order $d$ and dimensions $n_1, \ldots, n_d$. The elements of a tensor are given by its indices, ($i_1$,$i_2$,...,$i_d$), for which the MATLAB notation is used: 1 $\leq$ $i_k$ $\leq$ $n_k$. Even though tensors can be worked with in their mathematical form, it is easier to use a visual representation, as shown in Figure~\ref{fig:VisualTensorRep}. A tensor is illustrated as a core (circle) with a number of outgoing edges (lines) equal to its order. Each edge in such a diagram corresponds with a specific index.

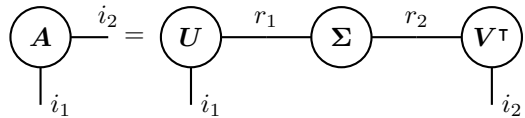
\begin{figure}[h!]
\centering
\begin{tikzpicture}
    \hspace{0.95cm}
    \draw[thick] (1,1) circle (0.4cm)   node {$\boldsymbol{A}$} ;
    \draw[thick] (1,0.6) -- (1,0.1)     node[right]{$i_1$};
    \draw[thick] (1.4,1) -- (1.9,1)     node[above]{$i_2$};   
    \node at (2.25, 1) {$=$};
    \draw[thick] (3,1) circle (0.4cm)   node {$\boldsymbol{U}$} ;
    \draw[thick] (3,0.6) -- (3,0.1)     node[right]{$i_1$};
    \draw[thick] (3.4,1) -- (4,1)       node[above]{$r_1$};  
    \draw[thick] (5,1) circle (0.4cm)   node {$\boldsymbol{\Sigma}$} ;
    \draw[thick] (4.6,1) -- (4,1);      
    \draw[thick] (5.4,1) -- (6,1)       node[above]{$r_2$}; 
    \draw[thick] (7,1) circle (0.4cm)   node {$\boldsymbol{V^{\intercal}}$} ;
    \draw[thick] (7,0.6) -- (7,0.1)     node[right]{$i_2$};
    \draw[thick] (6.6,1) -- (6,1);      
\end{tikzpicture}
    \caption{Tensor network representation of the SVD.}
    \label{fig:SVD_tensornetwork}
\end{figure}

Tensor networks are factorizations of tensors, analogous to matrix factorizations of matrices. In fact, it is a generalization of matrix decompositions to higher orders.  
An easy introduction is given through the singular value decomposition (SVD) of a matrix A
\begin{align*}
    \boldsymbol{A}(i_1,i_2) = \sum_{r_1}\sum_{r_2}  \boldsymbol{U}(i_1,r_1)\boldsymbol{\Sigma}(r_1,r_2) \boldsymbol{V}(i_2,r_2),
\end{align*}
into two orthogonal matrices $\boldsymbol{U}$,$\boldsymbol{V}$ and a diagonal matrix $\boldsymbol{\Sigma}$. It is one specific type of matrix factorization, and can therefore also be represented as a tensor network. The diagram for this matrix factorization is shown in Figure~\ref{fig:SVD_tensornetwork}. The interconnected edges represent the summations over the $r_1,r_2$ indices, also called index contractions. Generalized to higher orders, the shared edges between tensors represent tensor index contractions, which can be understood as a higher-order generalization of matrix multiplications.

\subsection{Tensor Train Vectors and Matrices}
\label{subsec:TensorTrains}
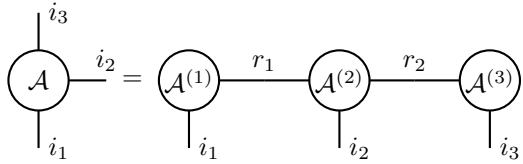
\begin{figure}[h!]
\centering
\begin{tikzpicture}
    \hspace{0.95cm}
    \draw[thick] (1,1) circle (0.4cm)   node {$\mathcal{A}$} ;
    \draw[thick] (1,1.4) -- (1,1.9)     node[right]{$i_3$}; 
    \draw[thick] (1,0.6) -- (1,0.1)     node[right]{$i_1$};
    \draw[thick] (1.4,1) -- (1.9,1)     node[above]{$i_2$};   
    \node at (2.25, 1) {$=$};
    \draw[thick] (3,1) circle (0.4cm)   node {$\mathcal{A}^{(1)}$} ;
    \draw[thick] (3,0.6) -- (3,0.1)     node[right]{$i_1$};
    \draw[thick] (3.4,1) -- (4,1)       node[above]{$r_1$};  
    \draw[thick] (5,1) circle (0.4cm)   node {$\mathcal{A}^{(2)}$} ;
    \draw[thick] (5,0.6) -- (5,0.1)     node[right]{$i_2$};
    \draw[thick] (4.6,1) -- (4,1);      
    \draw[thick] (5.4,1) -- (6,1)       node[above]{$r_2$}; 
    \draw[thick] (7,1) circle (0.4cm)   node {$\mathcal{A}^{(3)}$} ;
    \draw[thick] (7,0.6) -- (7,0.1)     node[right]{$i_3$};
    \draw[thick] (6.6,1) -- (6,1);      
\end{tikzpicture}
    \caption{Diagrammatic representation of a tensor train vector of a vector $\boldsymbol{a}$. First the vector was reshaped into a 3-dimensional tensor $\mathcal{A}(i_1,i_2,i_3)$.} 
    \label{fig:Tensor_Train}
\end{figure}

\begin{figure}[h!]
\centering
\begin{tikzpicture}
    \hspace{0.75cm}
    \draw[thick] (1,1) circle (0.4cm)   node {$\mathcal{A}$} ;
    \draw[thick] (0.775,1.305) -- (0.6,1.9)   node[above]{$i_2$}; 
    \draw[thick] (1,1.4) -- (1,1.9)     node[above]{$i_4$}; 
    \draw[thick] (1.225,1.305) -- (1.4,1.9)   node[above]{$i_6$};
    \draw[thick] (0.775,.695) -- (0.6,0.1)   node[below]{$i_1$}; 
    \draw[thick] (1,0.6) -- (1,0.1)     node[below]{$i_3$}; 
    \draw[thick] (1.225,0.695) -- (1.4,0.1)   node[below]{$i_5$};
    \node at (2, 1) {$=$};
    \draw[thick] (3,1) circle (0.4cm)   node {$\mathcal{A}^{(1)}$} ;
    \draw[thick] (3,1.4) -- (3,1.9)     node[right]{$i_2$}; 
    \draw[thick] (3,0.6) -- (3,0.1)     node[right]{$i_1$};
    \draw[thick] (3.4,1) -- (4,1)       node[above]{$r_1$};  
    \draw[thick] (5,1) circle (0.4cm)   node {$\mathcal{A}^{(2)}$} ;
    \draw[thick] (5,0.6) -- (5,0.1)     node[right]{$i_3$};
    \draw[thick] (4.6,1) -- (4,1);      
    \draw[thick] (5,1.4) -- (5,1.9)     node[right]{$i_4$}; 
    \draw[thick] (5.4,1) -- (6,1)       node[above]{$r_2$}; 
    \draw[thick] (7,1) circle (0.4cm)   node {$\mathcal{A}^{(3)}$} ;
    \draw[thick] (7,0.6) -- (7,0.1)     node[right]{$i_5$};
    \draw[thick] (6.6,1) -- (6,1);      
    \draw[thick] (7,1.4) -- (7,1.9)     node[right]{$i_6$}; 
\end{tikzpicture}
    \caption{Diagrammatic representation of a tensor train matrix decomposition. The matrix $\boldsymbol{A}$ was first reshaped into a 6-dimensional tensor $\mathcal{A}(i_1,\dots,i_6)$. The edges pointing downwards are row indices of the original matrix $\boldsymbol{A}$ and the edges pointing upwards are column indices of $\boldsymbol{A}$.}
    \label{fig:Tensor_Train_Matrix}
\end{figure}
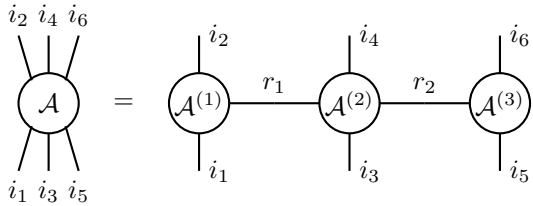

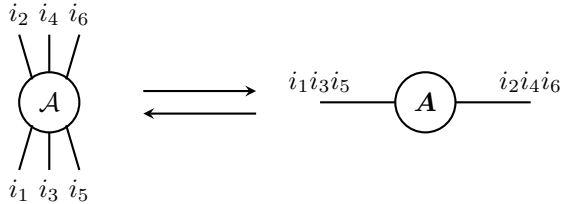
\begin{figure}[h!]
\centering
\begin{tikzpicture}
    \hspace{0.75cm}
    \draw[thick] (1,1) circle (0.4cm)   node {$\mathcal{A}$} ;
    \draw[thick] (0.775,1.305) -- (0.6,1.9)   node[above]{$i_2$}; 
    \draw[thick] (1,1.4) -- (1,1.9)     node[above]{$i_4$}; 
    \draw[thick] (1.225,1.305) -- (1.4,1.9)   node[above]{$i_6$};
    \draw[thick] (0.775,.695) -- (0.6,0.1)   node[below]{$i_1$}; 
    \draw[thick] (1,0.6) -- (1,0.1)     node[below]{$i_3$}; 
    \draw[thick] (1.225,0.695) -- (1.4,0.1)   node[below]{$i_5$};

    \tikzstyle{arrow} = [thick,->,>=stealth]
    \draw [arrow] (2.25,1.15) -- (3.75,1.15) ; 
     \draw [arrow] (3.75,0.85) -- (2.25,0.85) ; 
    \draw[thick] (6,1) circle (0.4cm)   node {$\boldsymbol{A}$} ;
    \draw[thick] (5.6,1) -- (4.6,1)       node[above]{$i_1 i_3i_5$}; 
    \draw[thick] (6.4,1) -- (7.4,1)       node[above]{$i_2i_4i_6$}; 
\end{tikzpicture}
    \caption{Arrow pointing to the right: a 6th-order tensor is reshaped into a matrix. Arrow pointing to the left: a matrix is reshaped into a 6th-order tensor.}
    \label{fig:Tensorization_matricization}
\end{figure}

In their high-dimensional form, tensors can be difficult to work with. Firstly, the computational and memory complexities $\mathcal{O}(n^d)$ are burdensome because of the exponential scaling with the order $d$. Secondly, concepts from linear algebra are not directly applicable, making analysis difficult. Tensor networks give more flexibility, as the tensor is factorized into a number of lower-dimensional tensor network components, also called cores. These factorizations can alleviate the adverse scaling and offer more insight. The CP and Tucker decompositions are two well-known tensor network structures, but have significant drawbacks, such as NP-hardness and poor scaling \cite{KoldaTensorDec}. The TT decomposition avoids these issues and can therefore be a more easy-to-use and robust format \cite{OseledetsTT,TTM_Oseledets}. As the name suggests, a TT decomposition factorizes a tensor into a chain of cores, which are linked through their TT-ranks $r_k$. In this paper, two types of tensor trains are used. A TT-vector TT$(\boldsymbol{a})$ to represent a vector $\boldsymbol{a}$, and a TT-matrix TT$(\boldsymbol{A})$ to represent a matrix $\boldsymbol{A}$. The diagrams of both a TT-vector and TT-matrix are shown in Figure~\ref{fig:Tensor_Train} and Figure~\ref{fig:Tensor_Train_Matrix}, respectively. Just like in Figure~\ref{fig:SVD_tensornetwork}, the connecting edges in the diagrams are indices that are summed over. To generate these TT structures, vectors ans matrices are first reshaped into high-order tensors as illustrated in Figure~\ref{fig:Tensorization_matricization} for the matrix case. Then, a low-rank TT decomposition is used to represent the obtained tensors. For example, a $11^3 \times 11^3$ matrix $A$ can be reshaped into a 6-way tensor with each dimension equal to 11. Both the row and column index are in this way split into 3 indices each, resulting in the 6-way tensor in Figure~\ref{fig:Tensor_Train_Matrix}. This 6-way tensor is then approximated by a low-rank tensor train decomposition. The TT-ranks $r_i$ signify the amount of correlation between the dimensions of a tensor, and determine the complexities of the representation. For small $r_i$, the TT format yields significant compression ratios as the storage complexities of a TT-vector and TT-matrix are $O(dnr^2)$ and $O(dn^2r^2)$, respectively.

Several different algorithms for converting a vector/matrix into their respective TT-form are available~\cite{holtz2012alternating,OseledetsTT,TTM_Oseledets,huber2017randomized}. In this article the TT-SVD is used as the algorithm is able to find a TT that satisfies guaranteed error bounds. The TT-SVD uses consecutive SVD operations to convert a tensor into a TT. Also the TT-rounding algorithm will be used in this article. The TT-rounding algorithm takes a given TT and truncates the TT-ranks, also through a sequence of consecutive SVD computations. There are two possible implementations of the TT-rounding algorithm: either an upper bound $\epsilon$ for the relative approximation error is provided or the TT-rank $r$ can be set to a pre-chosen value. For more details on these two algorithms we refer the reader to~\cite{OseledetsTT}.

\subsection{Tensor Train Operations}
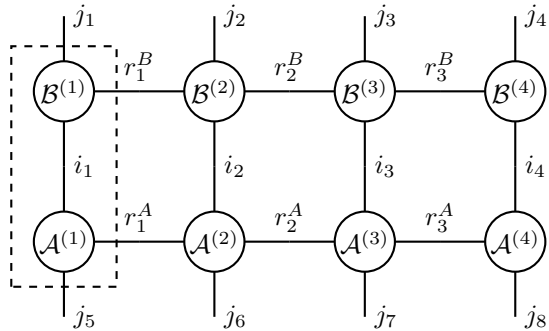
\begin{figure}[h!]
\centering
\begin{tikzpicture}
    \hspace{0.95cm}
    \draw[thick] (3,1) circle (0.4cm)   node {$\mathcal{B}^{(1)}$} ;
    \draw[thick] (3,1.4) -- (3,2)     node[right]{$j_1$};
    \draw[thick] (3,0.6) -- (3,0)     node[right]{$i_1$};
    \draw[thick] (3.4,1) -- (4,1)       node[above]{$r_1^B$};  
    \draw[thick] (5,1) circle (0.4cm)   node {$\mathcal{B}^{(2)}$} ;
    \draw[thick] (5,1.4) -- (5,2)     node[right]{$j_2$};
    \draw[thick] (5,0.6) -- (5,0)     node[right]{$i_2$};
    \draw[thick] (4.6,1) -- (4,1);      
    \draw[thick] (5.4,1) -- (6,1)       node[above]{$r_2^B$}; 
    \draw[thick] (7,1) circle (0.4cm)   node{$\mathcal{B}^{(3)}$} ;
    \draw[thick] (7,0.6) -- (7,0)     node[right]{$i_3$};
    \draw[thick] (7,1.4) -- (7,2)     node[right]{$j_3$};
    \draw[thick] (6.6,1) -- (6,1);      
    \draw[thick] (9,1) circle (0.4cm)   node{$\mathcal{B}^{(4)}$} ;
    \draw[thick] (9,0.6) -- (9,0)     node[right]{$i_4$};
    \draw[thick] (9,1.4) -- (9,2)     node[right]{$j_4$};
    \draw[thick] (7.4,1) -- (8,1)       node[above]{$r_3^B$};   
    \draw[thick] (8,1) -- (8.6,1);   
    
    \draw[thick] (3,-1) circle (0.4cm)   node {$\mathcal{A}^{(1)}$} ;
    \draw[thick] (3,-1.4) -- (3,-2)     node[right]{$j_5$};
    \draw[thick] (3,-0.6) -- (3,0) ;
    \draw[thick] (3.4,-1) -- (4,-1)       node[above]{$r_1^A$};  
    \draw[thick] (5,-1.4) -- (5,-2)     node[right]{$j_6$};
    \draw[thick] (5,-1) circle (0.4cm)   node {$\mathcal{A}^{(2)}$} ;
    \draw[thick] (5,-0.6) -- (5,0);
    \draw[thick] (4.6,-1) -- (4,-1);      
    \draw[thick] (5.4,-1) -- (6,-1)       node[above]{$r_2^A$}; 
    \draw[thick] (7,-1) circle (0.4cm)   node{$\mathcal{A}^{(3)}$} ;
    \draw[thick] (7,-1.4) -- (7,-2)     node[right]{$j_7$};
    \draw[thick] (7,-0.6) -- (7,0);
    \draw[thick] (6.6,-1) -- (6,-1);      
    \draw[thick] (9,-1) circle (0.4cm)   node{$\mathcal{A}^{(4)}$} ;
    \draw[thick] (9,-1.4) -- (9,-2)     node[right]{$j_8$};
    \draw[thick] (9,-0.6) -- (9,0);
    \draw[thick] (7.4,-1) -- (8,-1)       node[above]{$r_3^A$};   
    \draw[thick] (8,-1) -- (8.6,-1);  
    
    \draw[draw=black, thick, dashed] (2.3,-1.6) rectangle ++(1.4,3.2);
\end{tikzpicture}
    \caption{Diagrammatic representation of a matrix product in TT-matrix form. The resulting TT-matrix from this contraction TT$(\mathcal{C})$ consists of four cores, each with two free indices as given by the $j$ edges. The edges pointing downwards are row indices and the edges pointing upwards are column indices.}
    \label{fig:TensorTrainContraction}
\end{figure}

An advantage of the TT-vector and TT-matrix representations is that various linear algebra operations can be done explicitly in the compressed form. The two main operations that are required in the Kalman equations are matrix multiplication and matrix addition. Addition and subtraction in TT form is done by a concatenation procedure between respective cores \cite[p.~2308]{OseledetsTT}. Multiplication between a scalar and a TT is executed by multiplying the elements of one of the TT cores by the scalar. A matrix multiplication in TT-matrix form can be done via summations over indices between the two TT matrices. The procedure is illustrated for an example in Figure~\ref{fig:TensorTrainContraction}. The matrices $\boldsymbol{A},\boldsymbol{B}$ are both represented by TT matrices of 4 cores. The matrix product $\boldsymbol{A}\boldsymbol{B}$ is then computed in TT-matrix form by summing over the column indices of $\boldsymbol{B}$ and row indices of $\boldsymbol{A}$. These index summations ``merge" corresponding TT cores, as indicated by the dotted black rectangle in Figure~\ref{fig:TensorTrainContraction}. As a result of this ``merging", the TT-matrix ranks of the matrix product will be the product of the TT-matrix ranks of the individual matrices. This increase of the TT-matrix ranks motivates the use of the TT-rounding algorithm to truncate these ranks in order to keep computations feasible. The matrix product $\boldsymbol{A}\boldsymbol{B}$ in TT form is denoted as TT$(\boldsymbol{A}) \times \textrm{TT}(\boldsymbol{B})$.

\section{Tensor Network Kalman Filter}

The Kalman filter gives a simple iterative method for solving the dual problem, but the $n^d \times n^d$ covariance matrix $\boldsymbol{P}_k$ limits its feasibility. Its explicit representation demands an $\mathcal{O}(n^{2d})$ storage complexity. Fortunately, recasting the Kalman filter into a tensor network Kalman filter (TNKF) form avoids this problem, as has been done similarly in \cite{VolterraMIMOKimKF} and \cite{TNKF_LTI} for specific control applications. The reformulation is achieved through representing vectors by TT-vectors and matrices by TT-matrices, reducing the storage complexity to be linear in $d$ as indicated in Table~\ref{tab:TTKF_variables}. The TT-ranks ($r$) can be understood as hyperparameters that determine the accuracy-complexity trade-off. The The Kalman filter in TT form is summarized in Algorithm~\ref{alg:TNKF_algorithm}. 

\begin{table}[h!]
    \caption{Variables of the TNKF}
    \label{tab:TTKF_variables}
    \centering
    {\renewcommand{\arraystretch}{1.2}
    \begin{tabular}{|l|l|l|}
    \hline
     \textbf{Variables} & \textbf{TT variable} & \bfseries\makecell[l]{Storage \\ complexity}  \\ \hline
    $P_k$   & TT$(\boldsymbol{P}_k)$ & $\mathcal{O}(dn^2r^2)$ \\
    $\boldsymbol{c}_k$, $\boldsymbol{m}_k$, $\boldsymbol{k}_k$   &   TT$(\boldsymbol{c}_k)$, TT$(\boldsymbol{m}_k)$, TT$(\boldsymbol{k}_k)$ & $\mathcal{O}(dnr^2)$\\
    $y_k$, $v_k$, $s_k$, $w$  & Unchanged & $\mathcal{O}(1)$ \\ \hline 
    \end{tabular}
    }
\end{table}

\begin{algorithm}[h!]
\caption{Tensor Network Kalman Filter}
\label{alg:TNKF_algorithm}
\begin{algorithmic}
    \REQUIRE TT$(\boldsymbol{m}_0)$, TT$(\boldsymbol{P}_0)$, $\sigma_u^2$, $k\left(\boldsymbol{x}', \boldsymbol{x}\right)$, $\{\boldsymbol{x}_i,y_i\}_{i=1}^{N}$, $\gamma$, $\lambda$.
    \WHILE{Termination conditions not met} 
    \STATE   1. TT$(\boldsymbol{m}^{-}_{k}) = \textrm{TT}(\boldsymbol{m}_{k-1})$
    \STATE   2. TT$(\boldsymbol{P}^{-}_{k}) = \textrm{TT}(\boldsymbol{P}_{k-1})+\boldsymbol{Q}$
    \STATE   3. TT$(\boldsymbol{c}_k) =$ \text{TT-SVD} $(\boldsymbol{c}_k)$
    \STATE   4. $v_k$ = $y_k- \textrm{TT}(\mathcal{C}_k) \times
    \textrm{TT}(\boldsymbol{m}^{-}_{k}) $  
    \\    
    \STATE  5. $s_k$ =  TT$(\boldsymbol{c}_k) \times \textrm{TT}(\boldsymbol{P}^{-}_{k}) \times \textrm{TT}(\boldsymbol{c}_k)^{\intercal}$ + $\sigma_r^2$ 
    \STATE  6. TT$(\boldsymbol{k}_k)$ = $ \left(\textrm{TT}(\boldsymbol{P}_k^-) \times \textrm{TT}(\boldsymbol{c}_{k}) \right)\cdot s_k^{-1}$
    
    \STATE  7. TT$(\boldsymbol{m}_k)$ = TT$(\boldsymbol{m}_{k}^{-}) + \textrm{TT}(\boldsymbol{k}_k)\cdot v_k$

    \STATE  8. TT$(\boldsymbol{P}_k)$ = TT$(\boldsymbol{P}_{k}^{-}) +\textrm{TT}(\boldsymbol{k}_k) \times  \textrm{TT}(\boldsymbol{k}_k)^{\intercal} \cdot s_k$
    
    \ENDWHILE
\end{algorithmic}
\end{algorithm}

The initial mean and covariance matrix of the prior distribution $\mathcal{P}\left(\bar{\boldsymbol{\alpha}}\right)$ are constructed directly in TT form as the explicit construction of the covariance matrix can be impractical or even impossible for large datasets. In \cite{VolterraMIMOKimKF}, a more detailed explanation is given on how a rank-1 TT-vector and matrix can be constructed for the mean and covariance, respectively. In each iteration, a row of the kernel matrix is constructed with the training data and kernel function. This row is then reshaped into a tensor, after which it is decomposed to a low-rank TT-vector. The necessity of storing more than one row of the kernel matrix is thereby avoided, and permits application of very large datasets. 

Table~\ref{tab:compu_compl_tnkf} lists the computational complexity of each step of the TN Kalman filter a. The maximal TT-ranks for $\boldsymbol{P}_k,\boldsymbol{m}_k,\boldsymbol{k}_k$ are denoted $r_{\boldsymbol{P}},r_{\boldsymbol{m}},r_{\boldsymbol{k}}$, respectively. 
From the TNKF complexities, which includes the construction of the $k$-th row of the kernel matrix, it is clear that the TT-ranks play an important role in the complexity of the method. Especially the TT-ranks of TT$(\boldsymbol{c}_k)$ and TT$(\boldsymbol{P}_k)$ determine much of the algorithm's computational cost, as they are required almost every step, and impact the TT-ranks of the other variables. It is also evident that the construction of the TT-vector of $\boldsymbol{c}_k$ can be a dominating factor for large datasets.

An alternative for the TT-SVD algorithm with lower computational complexity would be to use an Alternating Linear Scheme (ALS)~\cite{holtz2012alternating} approach where the TT of the previous row could be used as an initial guess for the next row. Other alternatives with lower computational complexity are the randomized TT-SVD~\cite{huber2017randomized} and cross-approximation algorithm~\cite{oseledets2010tt}. In comparison to other low-rank approximation methods such as the Nystr\"om and FS-LSSVM method, the TNKF is advantageous if the data admits a low-TT-rank structure and when the spectrum of the kernel matrix decays slowly.

The TNKF is also advantageous compared to common iterative methods, such as conjugate gradient. These have computational complexities $\mathcal{O}(n^{2d}L)$, where $L$ is the number of iterations until convergence, and therefore suffer from the curse of dimensionality \cite{MatrixComputations}. As a result, iterative methods quickly become infeasible for large-scale datasets.

\newcommand\xrowht[2][0]{\addstackgap[.5\dimexpr#2\relax]{\vphantom{#1}}}

\begin{table}[h!]
    \caption{Computational complexities of the TNKF}
    \label{tab:compu_compl_tnkf}
    \centering
    \begin{tabular}{|l|l|l|}
    \hline \xrowht{6.5pt}
    \textbf{Step} & \textbf{Computational complexity} & \textbf{Explanation} \\ \hline \xrowht{6.5pt}
    1.  & $\mathcal{O}(1)$  & Prediction update TT($\boldsymbol{m}_k$)                        \\ \hline \xrowht{6.5pt}
    2.  & $\mathcal{O}(r_{\boldsymbol{P}}n^2)$   & Prediction update TT($\boldsymbol{P}_k$)   \\ \hline \xrowht{5.5pt}
    3a.  & $ \mathcal{O}(n^{d+2})$    & Construct kernel row   \\ \hline  \xrowht{6.5pt}   
    3b.  & $\mathcal{O}(1)$  & Reshape $\boldsymbol{c}_k$ to tensor  \\ \hline \xrowht{6.5pt}
    3c.  & $\mathcal{O}(dnr_{\boldsymbol{c}}^3)$ & {TT-SVD}$(\boldsymbol{c}_k)$                \\ \hline \xrowht{6.5pt}
    4.  & $\mathcal{O}(dnr_{\boldsymbol{c}}^2r_{\boldsymbol{m}}^2)$   & Compute $v_k$        \\ \hline \xrowht{6.5pt}
    5.  & $\mathcal{O}(dnr_{\boldsymbol{c}}^4r_{\boldsymbol{P}}^2)$  & Compute $s_k$              \\ \hline \xrowht{6.5pt}
    6a.  & $\mathcal{O}(dn^2r_{\boldsymbol{c}}^2r_{\boldsymbol{P}}^2)$ & Compute TT($\boldsymbol{k}_k$) \\ \hline \xrowht{6.5pt}
    6b.  & $\mathcal{O}(dnr_{\boldsymbol{k}}^2+dr_{\boldsymbol{k}}^4)$ & TT-Rounding TT($\boldsymbol{k}_k$)            \\ \hline \xrowht{6.5pt}
    7a.  & $\mathcal{O}(nr_{\boldsymbol{k}})$                         & Measurement update TT($\boldsymbol{m}_k$)            \\ \hline \xrowht{6.5pt}
    7b. & $\mathcal{O}(dnr_{\boldsymbol{m}}^2+dr_{\boldsymbol{m}}^4)$  & TT-Rounding TT($\boldsymbol{m}_k$)                 \\ \hline \xrowht{6.5pt}
    8a. & $\mathcal{O}(dn^2r_{\boldsymbol{k}}^4)$ & Measurement update TT($\boldsymbol{P}_k$)                    \\ \hline \xrowht{5.5pt}
    8b. & $\mathcal{O}(dnr_{\boldsymbol{p}}^2+dr_{\boldsymbol{P}}^4)$  & TT-Rounding TT($\boldsymbol{P}_k$)                     \\ \hline 
\end{tabular}
\end{table}

\subsection{Computation of Confidence Bounds}
\label{sec:ConfidenceBounds}
The output of Algorithm~\ref{alg:TNKF_algorithm} are the mean and covariance matrix of the the posterior distribution  $\mathcal{P}(\bar{\boldsymbol{\alpha}}| y_{1:n^d})=  \mathcal{N}(\textrm{TT}(\boldsymbol{m})$, $ \textrm{TT}(\boldsymbol{P}))$ in TT form, which can then be used for validation and testing. The dual model~\eqref{eq:dualmodel} can be implemented in TT form to keep working with lower computational and storage complexities. Consider $N$ test points $\{\boldsymbol{x}_i\}_{i=1}^{N}$. Similarly to Algorithm~\ref{alg:TNKF_algorithm}, rows of the kernel matrix based on the training and test data are constructed and transformed to a TT-vector TT$(\boldsymbol{c}_k)$. The prediction is then obtained by computing TT$(\boldsymbol{c}_k^T) \times$TT$(\boldsymbol{m})$. The variance $\sigma_y^2$ of the prediction is then TT$(\boldsymbol{c}^t_k) \times \textrm{TT}(\boldsymbol{P})\times\textrm{TT}(\boldsymbol{c}^t_k)^{\intercal}$+$\sigma_r^2$ and can be used to construct the $\pm\, 3\sigma_y$ (99.73\%) confidence bounds. The bounds describe how well the kernel regression or classification models fit the data. In Algorithm \ref{alg:TNKF_regression}, the TT test regression procedure is given. A similar procedure can be derived for classification. 

\begin{algorithm}[h!]
\caption{TNKF Regression - Prediction and computation of confidence bounds.}
\label{alg:TNKF_regression}
\begin{algorithmic}
    \REQUIRE TT$(\boldsymbol{m}_N)$, TT$(\boldsymbol{P}_N)$, $\sigma_u^2$, $k\left(\boldsymbol{x}', \boldsymbol{x}\right)$, training inputs $\{\boldsymbol{x}_i\}_{i=1}^{n^d}$, test inputs $\{\boldsymbol{x}^*_{i}\}_{i=1}^{N}$
    \FOR{$j$ = $1$ : $(N)$} 
    \STATE   1. TT$(\boldsymbol{c}_j)$ = TT-SVD$(\boldsymbol{c}_j)$. 
    \STATE   2. $\bar{y}_j$ = TT$(\boldsymbol{c}^T_j) \times \textrm{TT}(\boldsymbol{m}_N)$
    \STATE   3. $\sigma^2_j$ = TT$(\boldsymbol{c}^T_j) \times \textrm{TT}(\boldsymbol{P}_N) \times \textrm{TT}(\boldsymbol{c}_j)+\sigma_r^2$
    \ENDFOR
\end{algorithmic}
\end{algorithm}

\subsection{Practical Considerations}
Many different factorizations of the dimensions can be considered when converting vectors and matrices into the TT form. For example, each row of a $1000 \times 1000 $ kernel matrix could be ``tensorized'' into either a $10 \times 10 \times 10$ or $2 \times 5 \times 10 \times 10$ tensor, and so on. A general recommendation is to use the prime factorization since then we have decomposed the dimension into its smallest constituents. Using a prime factorization does however not uniquely determine how we can ``tensorize'' our vectors and matrices since the TT-ranks are known to depend on the order of the indices. For this reason it is also recommended to sort the dimensions. For the $1000 \times 1000$ kernel matrix, this means that each row would be reshaped into a $ 2 \times 2 \times 2 \times 5 \times 5 \times 5$ tensor.

Choosing the rounding for the TT ranks is non-trivial, as there is no way to determine their values a priori. Manual tuning and grid searches on subsets of the data can be used to determine suitable values. 

Early stopping criteria can be designed based on the norm-values of the covariance  $\boldsymbol{P}_k$. If the norm is small and has only slight value changes over the iterations, then there is high confidence in the solution in iteration $k$. Determining values for the covariance-based stopping criteria can be done by evaluating the primal error distributions and/or the residuals' variances, which are related to the dual weights by ${\alpha}_k \sim \mathcal{N}(0,\gamma^2 \sigma_{e}^2)$. Terminations criteria can also be based on other, such as relaxed Karush-Kuhn-Tucker (KKT) conditions.

One advantage of the TNKF has is that truncation-parameters allow it to be less sensitive to overfitting because the TT-ranks act as an implicit regularization. Truncations are easiest and most beneficial when the hyperparameters are small because this produces a rapidly decaying singular value spectrum for which small TT-ranks can be achieved in the TT-SVD. However, the selection of truncation and kernel parameters is interdependent. A balance needs to be found between them in order to fit the data well. A consequence of performing truncations is that the confidence region becomes wider, locally or globally, allowing for more uncertainties in the problem.

The TNKF performance is also impacted by the specified residual variances which determine the initial distributions. These variances influence the balance between generalization and overfitting, affecting both the width of the confidence bounds and the accuracy of the mean. In the supplied initial distribution, the mean does not impact regression. The first row of the dual problem is a KKT optimality condition ($\sum \alpha_i  = 0$) that resets the mean to zero. The initial covariance does impact the regression, but only if significant truncations are performed or early stopping is used in the TNKF, such that local variances do not converge to the $\sigma_r^2$. If no primal residual variance is known, the prior distribution can be chosen as $\boldsymbol{P_0} := \sigma_{r}^2 \boldsymbol{I}$. The dual residual covariance $\sigma_r^2$ determines the lower bound for the covariance in the Kalman equations, and can be used to supply an initial distribution. Alternatively, a user-specified initial distribution can also be supplied.

\section{Experiments}
In this section, the TNKF algorithm is applied to a number of large-scale regression and classification problems and compared to the FS-LSSVM and Nystr\"om methods. Table~\ref{tab:datasets_in_paper} gives an overview of the datasets, together with how many data points were used for training and validation and the dimension of the input space.

\begin{table}[h!]
    \caption{Considered datasets in this paper}
    \label{tab:datasets_in_paper}
    \centering
    \begin{tabular}{|l|l|l|l|}
        \hline
        \textbf{Datasets} & \bfseries\makecell[l]{\# Training\\samples} &  \bfseries\makecell[l]{\# Test\\ samples}  & \bfseries\makecell[l]{ Input \\ dimension $f$} \\ \hline \xrowht{6.5pt}
        Noisy sinc~\cite{LS-SVM} & $16384 = 2^{14}$ & $8096 = 2^{13}$ & 1 \\ \hline \xrowht{6.5pt}
        Two-spiral~\cite{LS-SVM}     & $1048576 = 2^{20}$ & $262144 = 2^{18}$ & 2 \\ \hline \xrowht{6.5pt}
        Adult~\cite{UCI_adult} & $ 19683 = 3^{9}$  & $ 2187 = 3^{7}$  & 16 $\xrightarrow{}$ 99  \\ \hline  \xrowht{6.5pt}
        MNIST~\cite{MNIST-data,MNISTcsv} & $59049 = 3^{10}$ & $6561 = 3^{8}$ & 784 \\ \hline
    \end{tabular}
\end{table}
Because of the differences between the considered low-rank approximation methods, a direct comparison is not straightforward. The following steps and assumptions are made to simplify the comparison.

\begin{itemize}
    \item The Nystr\"om method uses uniform random sampling to generate a subset of (S) data points. FS-LSSVM uses the Renyi-entropy criterion to generate a subset of (S) data points \cite{LS-SVM}. 
    \item The number of Renyi sampling iterations for FS-LSSVM is chosen the same as the number of iterations of the TNKF, equal to the number of data points.
    \item Multiple performance measures are shown for each method including different truncation parameters for the TNKF, subset sizes (S), and the number of eigenvectors (EV) in the Nystr\"om approximation.
    \item The same kernel hyperparameters are used to compare the methods. These were found by grid searches to determine the approximate orders, and final manual iterative tuning of the TNKF.
    \item For regression, the root-mean-square error (RMSE)
    \begin{align*}
        \text{RMSE} &=\sqrt{\sum_{i=1}^{N^t} \frac{\left({y}_{i}-\hat{y}_{i}\right)^{2}}{N}}
    \end{align*}
    and fit percentage based on the normalized root-mean-square error (NMRSE)
    \begin{align*}
        \text{Fit (NMRSE)} &= 100\cdot\left(1-\frac{\|\boldsymbol{y}-{\boldsymbol{\hat{y}}}\|}{\|\boldsymbol{y}-\operatorname{mean}(\boldsymbol{\hat{y}})\|}\right) \%
    \end{align*}
    are used as performance measures.
    \item For classification, the performance is given through the percentage of correctly assigned labels. The confidence (\%) is based on the total number of misclassifications by considering the ($\pm 3\sigma$) confidence bounds as decision models. 
    \begin{equation*}
     \text{Confidence} =  100 \cdot  \left( \frac{\# y_i \notin \left[ \hat{y}_i+3{\sigma}_{y_i},
     \hat{y}-3\sigma_{y_i} \right]}{N} \right) \%
    \end{equation*}
\end{itemize}

In the upcoming subsections, the following conventions are used: all datasets are centered, effectively removing the bias term and first column of the dual problem. This means that only the dual weights $\boldsymbol{\alpha}$ are learned. The bias term can simply be computed by calculation of the mean and it can be added to the final model if desired. In the performance tables, "NA" stands for "not applicable". This is when the algorithm is infeasible because of computational or memory requirements, or because no confidence bounds are generated. Multiple instances of the TNKF are tested for each dataset to analyze the effects of TT-rank truncations. Each TNKF instance has different truncation parameters, which are presented below the performance tables. The number of samples for Nystr\"om and FS-LSSVM is given through "\#S" after their names. 

All experiments were conducted in MATLAB - version R2019b, and performed on a 1aptop with an Intel 6-Core i7 processor running at 2.6 GHz and 16GB of RAM. The MATLAB code can be downloaded from \url{https://github.com/maximilianluc/LSSVM-TNKF}. For the FS-LSSVM and Nystr\"om methods the "ls-svmlab" toolbox was used, freely available at \url{https://www.esat.kuleuven.be/sista/lssvmlab/}. 

The hyperparameters and truncation parameters were found through iterative grid searches with increasingly fine grid spacing. Currently there are no methods to determine these parameter values beforehand. Additionally, the hyperparameters and truncation parameters values are interdependent.  

\subsection{Regression Problem 1: Noisy Sinc Function}
To find an approximation of $\boldsymbol{\alpha}$, the TNKF needs to be initialized. The sinc function is corrupted with noise distributed as $ e \sim \mathcal{N}(0,0.1^2)$. It is assumed that this is known, and $r$ is chosen equal to this noise. The RBF kernel function is used and hyperparameter values chosen as: \mbox{($\gamma = 0.005$, $\sigma_{RBF}^2 = 0.005$)}. The initial distribution of $\boldsymbol{\alpha}$ is given as $\mathcal{N}(\boldsymbol{0}, diag[0.1^2 \cdot 0.005^2])$ according to the relation $\boldsymbol{P_0} :=\gamma^2 \sigma_{e}^2 \boldsymbol{I}$, and supplied in TT form, emphasizing high confidence in the initial solution.  

\begin{table}[h!]
    \caption{Test performance of the approximation methods on the noisy sinc function  (RBF kernel, $\gamma = 0.005$, $\sigma_{RBF}^2 = 0.005$). }
   \centering
   \subfloat[Performance values \label{tab:perf_noisy_sinc_a}]{
    \centering
    { 
    \renewcommand{\arraystretch}{1.2}
    \begin{tabular}{|l|l|l|l|l|}
        \hline
        \textbf{Method} & \bfseries\makecell[l]{Training \\ RMSE} & \bfseries\makecell[l]{Test \\ RMSE} & \bfseries\makecell[l]{Test \\ Fit \% }   \\ \hline
        TNKF$^{\hspace{0.05cm}a}$  & 0.160 & 0.162 & 69.5 \\ \hline
        TNKF$^{\hspace{0.05cm}b}$ & 0.104 & 0.104 &  75.4 \\ \hline
        FS-LSSVM 20 S & 0.261 & 0.260 & 25.5\\ \hline
        FS-LSSVM 100 S & 0.100 & 0.102 & 75.3 \\ \hline
        FS-LSSVM 500 S & $\boldsymbol{0.099}$ & $\boldsymbol{0.099}$ & $\boldsymbol{76.0}$ \\ \hline
        Nystr\"om $2^{14}$ S, 10 EV  & 0.788 & 0.786 & 31.6 \\ \hline
        Nystr\"om $2^{14}$ S, 50 EV  & 0.146 & 0.148 & 60.2 \\ \hline
        Nystr\"om $2^{13}$ S, 50 EV  & 0.194 & 0.180 & 34.7 \\ \hline
        Nystr\"om $2^{12}$ S, 75 EV  & 0.197 & 0.185 & 28.9 \\ \hline
        
    \end{tabular}}}\\
   \subfloat[Truncation values for the TNKF \label{tab:trunc_noisy_sinc}]{
     \centering
        { 
        \renewcommand{\arraystretch}{1.5}
        \begin{tabular}{|llllll|}
        \hline
        & $\epsilon_{\boldsymbol{m}}$ &$\epsilon_{\boldsymbol{c}}$ &$\epsilon_{\boldsymbol{P}}$ & $\epsilon_{\boldsymbol{k}}$ & $\epsilon_{\boldsymbol{y}^t}$ \\ \hline
            $^{a.}$ \footnotesize & 0.005 & 0.1 & 0.02 & 0.005  & 0.001  \\
            $^{b.}$ \footnotesize & 0 & 0.001 & 0.0005 & 0.2  & 0.001\\
            \hline
        \end{tabular}}
   }
\end{table}

\begin{figure}[h!]
    \centering
    \includegraphics[scale=0.25]{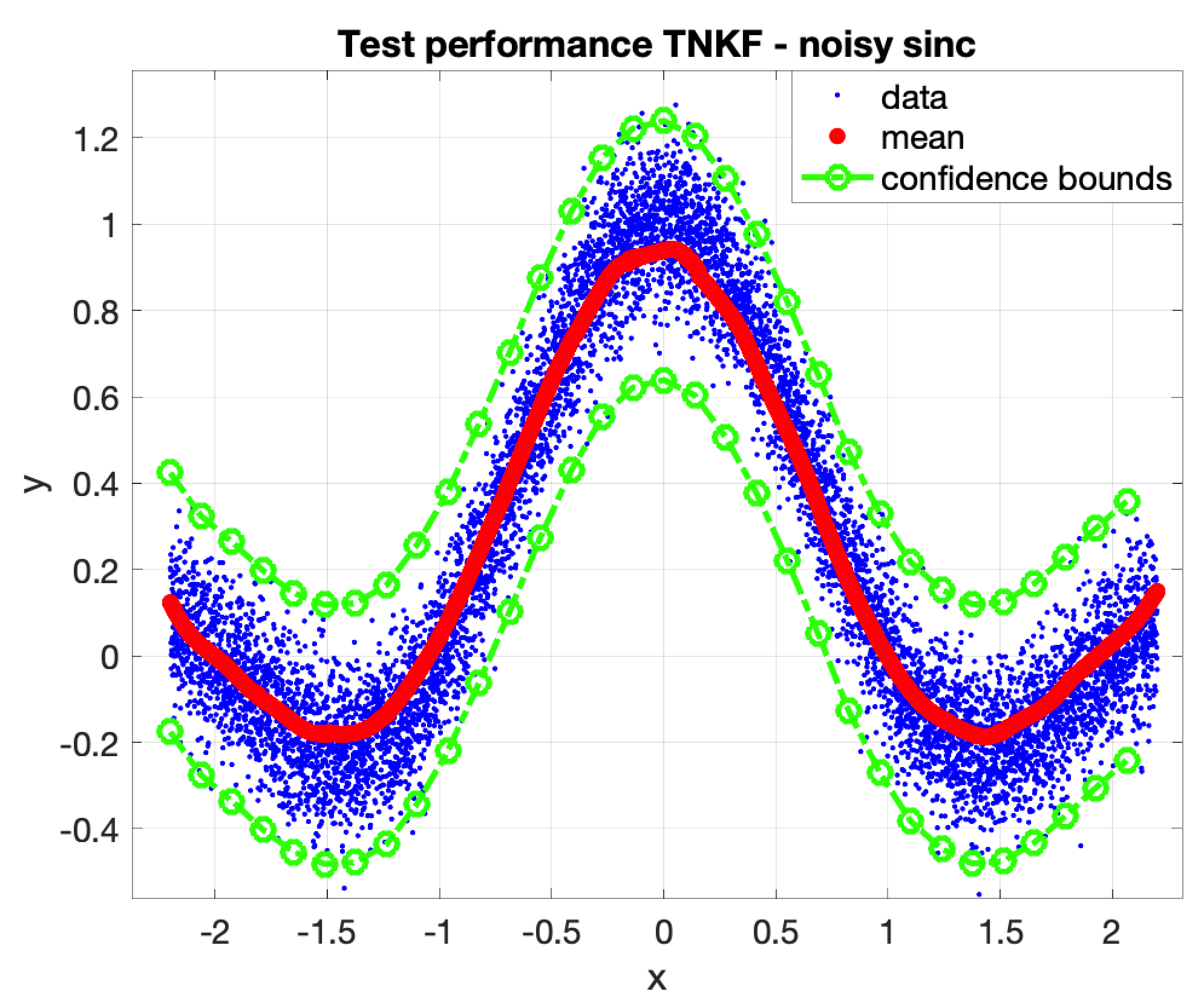}
    \caption{Regression performance of TNKF$^{b}$ with an initial distribution of $\boldsymbol{\alpha} \sim \mathcal{N}(0,\gamma^2\sigma_{e}^2\boldsymbol{I}) \sim \mathcal{N}(0,0.005^2 \cdot 0.1^2 \boldsymbol{I})$. The residual distribution is set equal to the noise of the underlying data, $r\sim\mathcal{N}(0,0.1^2)$.}
    \label{fig:TNKF_c_initDist}
\end{figure}

The performance of the TNKF is dependent on its TT-ranks, determined by the truncations and selected kernel hyperparameters, as given in Table VI-(a). If both are appropriately specified, the TNKF can obtain near-optimal RMSE ($0.1$) and fit values, given by the noise's standard deviation. The performance is poor when the truncations are specified too large, proven by TNKF$^{\hspace{0.05cm}a}$, given in Table VI-(b). The truncation of the kernel rows in TNKF$^{\hspace{0.05cm}a}$ results in too much information loss in comparison to TNKF$^{\hspace{0.05cm}b}$. Crucial interrelationships in the data are lost, therefore the prediction power of the TNKF is reduced.
The FS-LSSVM also manages to realize near-optimal results. Additionally, through iterative tuning of the sample sizes to analyze the complexity-accuracy trade-off, it is apparent that very few are needed to generate an accurate prediction for the LS-SVM. The method may therefore be preferable over the TNKF in terms of complexity. Similarly to the TNKF method, FS-LSSVM is less sensitive to the chosen hyperparameters. With the Renyi-sampling criterion, this method can select the most informative columns to acquire a nonlinear mapping function. The Renyi-sampling criterion makes it unnecessary to use larger subsets as the increase in performance is minimal compared to the increase in complexity. The Nystr\"om method is much more sensitive to which and the number of columns sampled. As presented in Table VI-(a), it does not match the performances of the TNKF and FS-LSSVM. Unlike the TNKF and FS-LSSVM methods, it is not able to select the most informative columns to overcome its susceptibility to noise. As a result, and due to the noise, the eigendecomposition is not accurate. In Figure~\ref{fig:TNKF_c_initDist}, the initial distribution of $\boldsymbol{\alpha}$ in TNKF$^{b}$ is set equal to the noise distribution, which results in smooth confidence bounds. 

\begin{figure}[h!]
    \centering
    \includegraphics[scale=0.25]{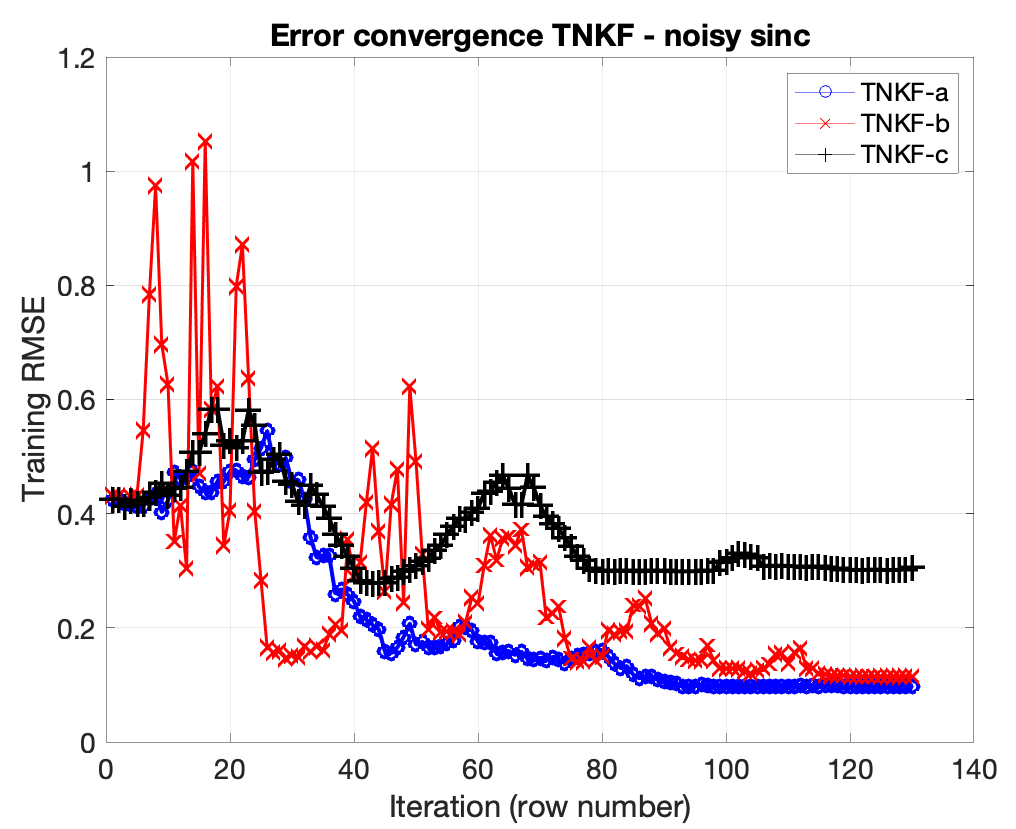}
    \caption{Training RMSE of three TNKF implementations on a small noisy sinc problem. The TNKF truncations, hyperparameters, initialization can have significant effect on the accuracy. The TNKF truncation values are provided in Table~\ref{tab:TNKF_errorconv}.}
    \label{fig:TNKF_errorconv}
\end{figure}

\begin{table}[h!]
    \caption{Truncation values for the TNKF error plots Fig.~\ref{fig:TNKF_errorconv}.}
    \label{tab:TNKF_errorconv}
    \centering
    { 
    \renewcommand{\arraystretch}{1.5}
    \begin{tabular}{|llllll|}
        \hline
        & $\epsilon_{\boldsymbol{m}}$ &$\epsilon_{\boldsymbol{c}}$ &$\epsilon_{\boldsymbol{P}}$ & $\epsilon_{\boldsymbol{k}}$ & $\epsilon_{\boldsymbol{y}^t}$ \\ \hline
            $^{a.}$ \footnotesize & 0.001 & 0.001 & 0.001 & 0.001  & 0.001  \\
            $^{b.}$ \footnotesize & 0.001 & 0.5 & 0.001 & 0.001  & 0.001 \\
            $^{c.}$ \footnotesize & 0.15 & 0.001 & 0.001 & 0.001  & 0.001 \\
            \hline
        \end{tabular}}
\end{table}

In Figure~\ref{fig:TNKF_errorconv}, the RMSEs of three different TNKFs over the iterations are plotted for a small scale noisy sinc problem. The respective truncation values are given in Table~\ref{tab:TNKF_errorconv}. Convergence of the TNKF depends significantly on the specified truncation values. For TNKF$^{c}$, the truncation of TT($\boldsymbol{m}_l$) prohibits convergence to the noise RMSE. The $\boldsymbol{\alpha}$ approximation does not accurately describe the dataset. The TNKF$^{b}$ convergence plot oscillates much more than for TNKF$^{a}$ because TT($\boldsymbol{c}_l$) truncations can make the training data produce less informative. TT($\boldsymbol{c}_l$) has a crucial role in whether the TNKF converges. Also, for TNKF$^{a}$ and TNKF$^{b}$, early stopping can be performed with small or minimal consequences to the test RMSE if tuned properly.

\subsection{Classification Problem 1: Two-spiral}

A large-scale version of the two-spiral problem is considered, a well-known and difficult machine learning problem. The task is to classify two identical, but phase shifted, spirals from each other.  The data is separable, but a highly nonlinear decision boundary has to be found \cite{LS-SVM}. The RBF kernel is used, and the hyperparameters are: \mbox{($\gamma = 0.05$, $\sigma_{RBF}^2 = 5\cdot10^{-8}$)}. The initial distribution for the weights is chosen as, \mbox{$\boldsymbol{\alpha}$ $\sim$  $\mathcal{N}(\boldsymbol{0}, \sigma_r^2 \boldsymbol{I}_{N})$}, with $\sigma_{r}^2 = 1\cdot10^{-5}$. The data adheres to \mbox{TT-rank-1} structures for which the memory and computational difficulties only scale with $n$ and $d$. The test dataset was generated by selecting every fourth point of the two spirals for training.
\begin{figure}[h!]
    \centering
    \includegraphics[scale=0.25]{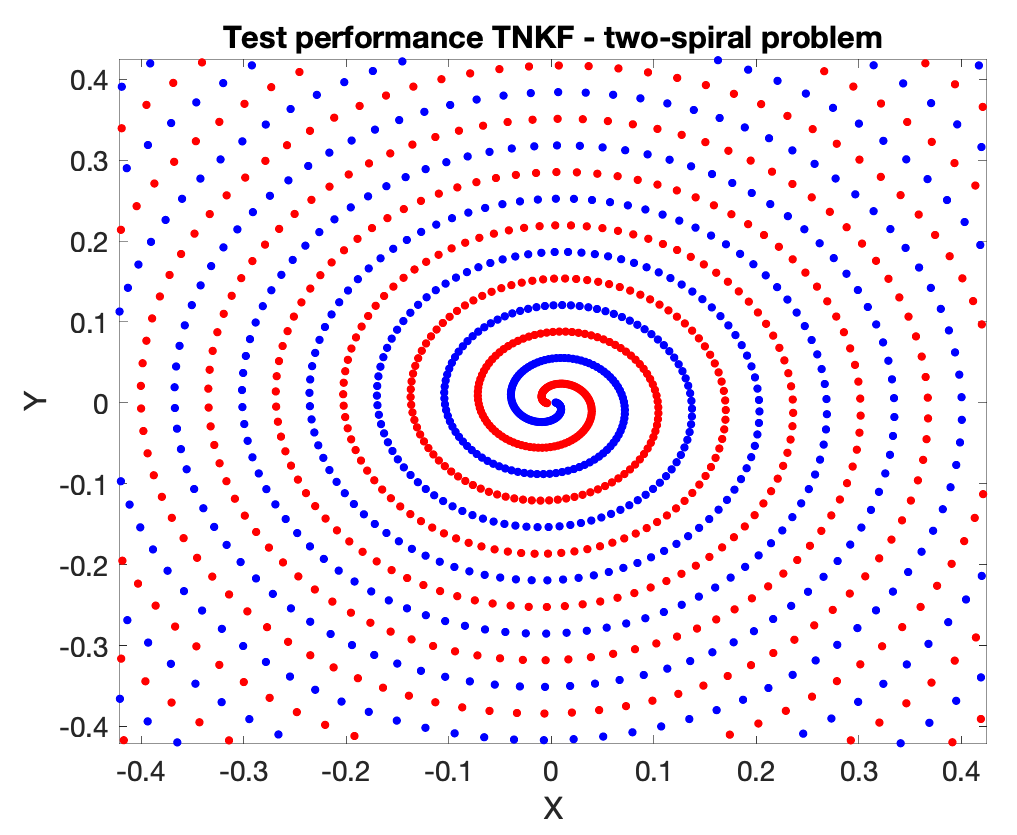}
    \caption{The two classified spirals for the test dataset TNKF$^{\hspace{0.05cm}a}$. A subset is shown for clarity. }
    \label{fig:two_sprial_performance}
\end{figure}

\begin{table}[h!]
    \label{tab:two_spiral_problem}
    \caption{Test performance of the approximation methods on the large-scale two-spiral problem  (RBF kernel, \mbox{$\gamma = 0.05$},   $\sigma_{RBF}^2 = 5\cdot10^{-8}$)}
   \centering
   \subfloat[Performance values \label{tab:twospiral_perf}]{
    \centering
    { 
    \renewcommand{\arraystretch}{1.25}
    \begin{tabular}{|l|l|l|}
        \hline
        \textbf{Method} &  \bfseries\makecell[l]{Correctly \\Labeled \%} & \textbf{Confidence \%}  \\ \hline
        TNKF$^{\hspace{0.05cm}a}$ & $\boldsymbol{100}$ & $\boldsymbol{100}$ \\ \hline
        FS-LSSVM & NA & NA \\ \hline
        Nystr\"om $2^{15}$ S, 250EV & 50.8 & NA  \\ \hline
        Nystr\"om $2^{15}$ S, 500EV & 50.7 & NA  \\ \hline
        Nystr\"om $2^{16}$ S, 250EV & NA & NA  \\ \hline
    \end{tabular}}}\\
   \subfloat[Truncation values for the TNKF \label{tab:twospiral_trunc}]{
     \centering
        { 
        \renewcommand{\arraystretch}{1.5}
        \begin{tabular}{|llllll|}
        \hline
       & $r_{\boldsymbol{m}}$ & $r_{\boldsymbol{c}}$ & $r_{\boldsymbol{P}}$ & $r_{\boldsymbol{k}}$ & $r_{\boldsymbol{y}^t}$ \\ \hline
        $^{a.}$ \footnotesize & 1 & 1 & 1 & 1 & 1 \\ \hline
        \end{tabular}}
   }
\end{table}

The TNKF instances manage to distinguish the two spirals perfectly, as given in Table VII and displayed in Figure~\ref{fig:two_sprial_performance}. Additionally, confidence bounds of the decision function are also found. Because of the small TT-ranks the TNKF is computationally advantageous, requiring no explicit storage of a large matrix. The FS-LSSVM classifier could only label a subset of the data therefore it is assigned "NA". Due to the size of the dataset and the small kernel hyperparameter it was impossible to generate a nonlinear mapping function
that generalizes well over the entire dataset. The Nystr\"om method also performed poorly, as the number of samples for a good estimation exceed RAM capabilities. In the case that fewer samples are chosen, the method also results in low labeling power. The large-scale two-spiral problem demonstrates that the TNKF is especially suitable in cases where kernel eigenspectrum decays slowly. The TNKF can incorporate the most important information from the dataset in an implicit way, not requiring a sampling procedure.

\subsection{Classification Problem 2: Adult Dataset}

In this experiment the goal is to classify whether a person makes over \$50k a year based on 14 features, both numerical and categorical \cite{UCI_adult}. By converting the categorical features to numerical values by one-hot-encoding, a total of 99 features is obtained. The RBF kernel is used, and the chosen hyperparameters are: ($\gamma = 0.0015$, $\sigma_{RBF}^2 = 0.5$). TNKF$^{\hspace{0.05cm}a}$ and TNKF$^{\hspace{0.05cm}b}$ are trained on the entire dataset. TNKF$^{\hspace{0.05cm}c}$  uses early stopping, and therefore witnesses only a subset of the data. The three TNKFs are implemented with initial distributions, \mbox{$\boldsymbol{\alpha}$ $\sim$ $\mathcal{N}(\boldsymbol{0},\boldsymbol{I}_N)$}. It is assumed that the residual is irrelevant to the consensus data, therefore it is chosen very small $r$ $\sim$ $\mathcal{N}(0,10^{-8})$. 

\begin{table}[h!]
    \label{tab:Adult_dataset}
    \caption{Test performance of the approximation methods on the Adult dataset  ($\gamma = 0.0015$, $\sigma_{RBF}^2 = 0.5$)}
   \centering
   \hspace{2cm}
   \subfloat[Performance values \label{tab:adult_perf}]{
    \centering
    { 
    \renewcommand{\arraystretch}{1.2}
    \begin{tabular}{|l|l|l|}
        \hline
        \textbf{Method} &  \bfseries\makecell[l]{Correctly \\Labeled \%}  & \textbf{Confidence \%}  \\ \hline
        TNKF$^{\hspace{0.05cm}a}$  & 83 & 98.2 \\ \hline
        TNKF$^{\hspace{0.05cm}b}$  & $\boldsymbol{84.5}$ & $\boldsymbol{99.2}$  \\ \hline
        TNKF$^{\hspace{0.05cm}c}$  & 81.4 & 96     \\ \hline
        FS-LSSVM 50 S    & 44.3 & NA  \\ \hline
        FS-LSSVM 500 S   & 73.2 & NA   \\ \hline
        FS-LSSVM 800 S  & 76.2 & NA  \\ \hline 
        Nystr\"om $3^6$, 50 EV  & 81.2 & NA \\ \hline
        Nystr\"om $3^7$, 25 EV  & 82.7 & NA \\ \hline
        Nystr\"om $3^9$, 25 EV  & 84.4 & NA \\ \hline

    \end{tabular}}}
   \newline
   \subfloat[Truncation values for the TNKF \label{tab:adult_trunc}]{
     \centering
        { 
        \renewcommand{\arraystretch}{1.5}        \begin{tabular}{|llllll|}
        \hline
        & $\epsilon_{\boldsymbol{m}}$ & $\boldsymbol{c}$ &$\epsilon_{P}$ &$ \boldsymbol{k}$ & $\epsilon_{\boldsymbol{y}^t}$ \\ \hline
        $^{a.}$ \footnotesize & 0.001 & $r_{\boldsymbol{c}}$=6 & 0.005  & $\epsilon_{\boldsymbol{k}}$=0.005 & 0.001 \\
        $^{b.}$ \footnotesize & 0.001 & $\epsilon_c$=0.01 & 0.003 & $\epsilon_{\boldsymbol{k}}$=0.003 & 0.001 \\
        $^{c.}$ \footnotesize & 0.001 & $r_c$=30 & 0.001 & $r_k$=6 &  0.001\\\hline
        \end{tabular}}}
\end{table}

In Table X-(a) the performances of the TNKF are given. The TNKF is capable of achieving performances equal to that found in the literature for SVM's and LS-SVMs \cite{LS-SVM, PerformanceAdultSVMs}. The data adheres to small TT-ranks, therefore severe truncations are unnecessary. TNKF$^{\hspace{0.05cm}c}$, which employed early stopping and a forgetting factor, managed to reach decent labeling power. Below the parameters for TNKF$^{\hspace{0.05cm}c}$ are described:

\begin{itemize}
    \item Forgetting factor $\lambda = (1.9975)^{-1}$,
    \item $||$TT$(\boldsymbol{P}_k)||_F$ smaller than $1\cdot10^{-5}$,
    \item The change in $||$TT$(\boldsymbol{P}_k)||_F$ at most $-5\cdot10^{-3}$ for at least 5 iterations.
\end{itemize}

Implementation of early stopping required only 4200 rows to be iterated over. This reduced training time by around 80\%, but yielded a lower classification accuracy. The chosen kernel hyperparameter ($\sigma_{RBF}^2$) generates a kernel matrix that can easily be approximated with a low-rank matrix structure. The Nystr\"om method is particularly suitable for this, and achieves very high performance values with few sampled columns. The nonlinear mapping function, however, is difficult to approximate. FS-LSSVM requires many support vectors and is computationally the most demanding in this problem.

\subsection{Classification Problem 3: binary MNIST}

For the MNIST digit recognition dataset a binary classification problem is considered in this paper. The digits are labeled in two groups, $1:=\{0,...,4\}$ and $-1:=\{5,...,9\}$, classified based on the 784 normalized pixels per image. A classifier is trained with $3^{10}$ images, and tested on $3^8$ images. No noise is considered to act on the data. The RBF kernel is used with a hyperparameter the selection \mbox{($\gamma = 0.05$, $\sigma_{RBF}^2 = 5$)}. The initial condition is chosen $\boldsymbol{\alpha} \sim \mathcal{N}(\boldsymbol{0},\boldsymbol{I}_N)$. The performances of the low-rank approximation methods on the binary MNIST classification problem are presented in Table XI.

\begin{table}[h!]
    \label{tab:MNIST_dataset}
    \caption{Test performance of the approximation methods on the MNIST dataset  ($\gamma = 0.05$, $\sigma_{RBF}^2 = 5$)}
    \centering
   \hspace{2cm}
   \subfloat[Performance values \label{tab:mnist_perf}]{
     {
    \renewcommand{\arraystretch}{1.2}
    \begin{tabular}{|l|l|l|}
        \hline
        \textbf{Method} &  \bfseries\makecell[l]{Correctly \\Labeled \%}  & \textbf{Confidence \%}  \\ \hline
        TNKF$^{\hspace{0.05cm}a}$ & $\boldsymbol{97.6}$ &  $\boldsymbol{99.2}$  \\ \hline
        FS-LSSVM 50 S & 61.4 & NA  \\ \hline
        FS-LSSVM 250 S & 75.3 & NA   \\ \hline
        FS-LSSVM 1000 S &  92.8 & NA  \\ \hline 
        Nystr\"om $3^4$ S, 10 EV  & 82.4 & NA \\ \hline
        Nystr\"om $3^7$ S, 50 EV  & 93.7 & NA \\ \hline
        Nystr\"om $3^9$ S, 25 EV  & 96.9 & NA \\ \hline
    \end{tabular}}}
   \newline
   \subfloat[Truncation values for the TNKF \label{tab:mnist_trunc}]{
        { 
        \renewcommand{\arraystretch}{1.5}
        
        \begin{tabular}{|llllll|}
        \hline
        & $r_{\boldsymbol{m}}$ & $r_{\boldsymbol{c}}$ & $r_{\boldsymbol{P}}$ & $r_{\boldsymbol{k}}$ & $r_{\boldsymbol{y}^t}$ \\ \hline
        $^{a.}$ \footnotesize & 4 & 1 & 1 & 2 & 2 \\ \hline
        \end{tabular}}}
\end{table}

The TNKF achieves high classification and confidence percentages, and is well-suited for the MNIST dataset because the images conform to small TT ranks. As a result of the low TT ranks, the memory and computational difficulty are small, even though not as favorable as the Nystr\"om method. The FS-LSSVM requires many support vectors and therefore demands much computational effort, nevertheless can reach high labeling power. The Nystr\"om approximations perform well with very few samples. Of the three low-rank approximation methods, it is computationally the lightest to apply and yields very high accuracies.

\section{Conclusion}

This paper presents a recursive Bayesian filtering framework with the tensor network Kalman filter to solve large-scale LS-SVM dual problems. To the best of our knowledge, it is the first non-sampling based algorithm for large-scale LS-SVM dual problems. The recursive Bayesian framework allows early stopping and yields confidence bounds on the predictions. Current low-rank approximation methods lack such properties. Also, the curse of dimensionality is avoided in this work because the method is iterative and uses a tensor train representation. The TNKF is especially advantageous in situations with high nonlinearities, when the kernel data conforms to a low-TT-rank structure, or when many samples are required for alternative low-rank methods. High accuracies are attained on all considered problems. 

In further research, the TNKF computational complexities can be further reduced. The algorithm can easily be extended to a parallel implementation and a batch framework to reduce training times. Additionally, directly constructing the kernel rows in TT form can save a significant portion of the computational complexity for extremely large problems and problems with many features. The tuning of the interdependent truncation parameters and kernel hyperparameters requires further investigation. Inference methods can be developed to determine these parameters and would save much tuning work, currently done by exhaustive grid searches. Lastly, the TNKF can be readily adopted to other kernel methods and fields involved with solving large-scale linear systems.

\section*{Acknowledgment}

Johan Suykens acknowledges support from EU: The research leading to these results has received funding from the European Research Council under the European Union’s Horizon 2020 research and innovation program/ERC Advanced Grant E-DUALITY (787960). This paper reflects only the authors’ views and the Union is not liable for any use that may be made of the contained information. Research Council KUL: C14/18/068 Flemish Government: FWO: projects: GOA4917N, Impulsfonds AI: VR 2019 2203 DOC.0318/1QUATER, Ford KU Leuven Research Alliance Project KUL0076, ICT 48 TAILOR, Leuven.AI Institute.

\clearpage
\bibliography{bibtex/bib/Bibliography.bib}{}
\bibliographystyle{unsrt.bst}

\end{document}